\pdfoutput=1
\documentclass[11pt,a4paper]{article}
\usepackage[utf8]{inputenc}
\usepackage[hyperref]{emnlp2020}
\usepackage{times}
\usepackage{latexsym}
\usepackage{svg}

\usepackage{CJKutf8}

\usepackage{microtype}
\usepackage{graphicx}
\usepackage{subfigure}
\usepackage{booktabs} 
\usepackage{adjustbox}
\usepackage{enumitem}
\usepackage{multirow, multicol}

\usepackage[T1]{fontenc}

\usepackage{microtype}
\usepackage{graphicx}
\usepackage{subfigure}
\usepackage{booktabs} 

\usepackage{amsmath,amsfonts,bm}

\def\eqref#1{equation~\ref{#1}}

\def\1{\bm{1}}

\def\mW{{\bm{W}}}

\DeclareMathAlphabet{\mathsfit}{\encodingdefault}{\sfdefault}{m}{sl}
\SetMathAlphabet{\mathsfit}{bold}{\encodingdefault}{\sfdefault}{bx}{n}

\newcommand{\softmax}{\mathrm{softmax}}

\usepackage{CJKutf8}

\usepackage{url}
\usepackage{graphicx}
\usepackage{booktabs}
\usepackage{multirow, multicol}
\usepackage{diagbox}

\usepackage{tabularx}
\usepackage{xspace}
\usepackage{amsmath}
\usepackage{float}
\usepackage{xcolor}
\usepackage{tablefootnote}

\newcommand{\experimentI}[0]{Bootstrap\xspace}
\newcommand{\experimentII}[0]{Bootstrap (+English)\xspace}

\newcommand{\experimentVdirect}[0]{BackTranslation\xspace}

\aclfinalcopy 
\def\aclpaperid{3597} 
\title{Localizing Open-Ontology QA Semantic Parsers \\ in a Day Using Machine Translation}

\author{Mehrad Moradshahi 
  \quad
  Giovanni Campagna
  \quad
  Sina J. Semnani 
  \quad
  Silei Xu
  \quad
  Monica S. Lam \\
  Computer Science Department \\
  Stanford University \\
  Stanford, CA, USA \\
  \texttt{\{mehrad,gcampagn,sinaj,silei,lam\}@cs.stanford.edu} \\}

\date{}
\begin{document}
\maketitle

\begin{abstract}

We propose Semantic Parser Localizer (SPL), a toolkit that leverages Neural Machine Translation (NMT) systems to localize a semantic parser for a new language. Our methodology is to (1) generate training data automatically in the target language by augmenting machine-translated datasets with local entities scraped from public websites, (2) add a few-shot boost of human-translated sentences and train a novel XLMR-LSTM semantic parser, and (3) test the model on natural utterances curated using human translators.


We assess the effectiveness of our approach by extending the current capabilities of Schema2QA, a system for English Question Answering (QA) on the open web, to 10 new languages for the restaurants and hotels domains. 
Our models achieve an overall test accuracy ranging between 61\% and 69\% for the hotels domain and between 64\% and 78\% for restaurants domain, which compares favorably to 69\% and 80\% obtained for English parser trained on gold English data and a few examples from validation set. 
We show our approach outperforms the previous state-of-the-art methodology by more than 30\% for hotels and 40\% for restaurants with localized ontologies for the subset of languages tested. 

Our methodology enables any software developer to add a new language capability to a QA system for a new domain, leveraging machine translation, in less than 24 hours. Our code is released open-source.\footnote{\footnotesize{\url{https://github.com/stanford-oval/SPL}}}

\end{abstract}

\section{Introduction}
\label{intro2}

Localization is an important step in software or website development for reaching an international audience in their native language. Localization is usually done through professional services that can translate text strings quickly into a wide variety of languages. As conversational agents are increasingly used as the new interface, how do we localize them to other languages efficiently? 



The focus of this paper is on question answering systems that use \textit{semantic parsing}, where natural language is translated into a formal, executable representation (such as SQL). Semantic parsing typically requires a large amount of training data, which must be annotated by an expert familiar with both the natural language of interest and the formal language.
The cost of acquiring such a dataset is prohibitively expensive.  

\begin{figure}[tb]
\centering
\includegraphics[width=0.9\linewidth]{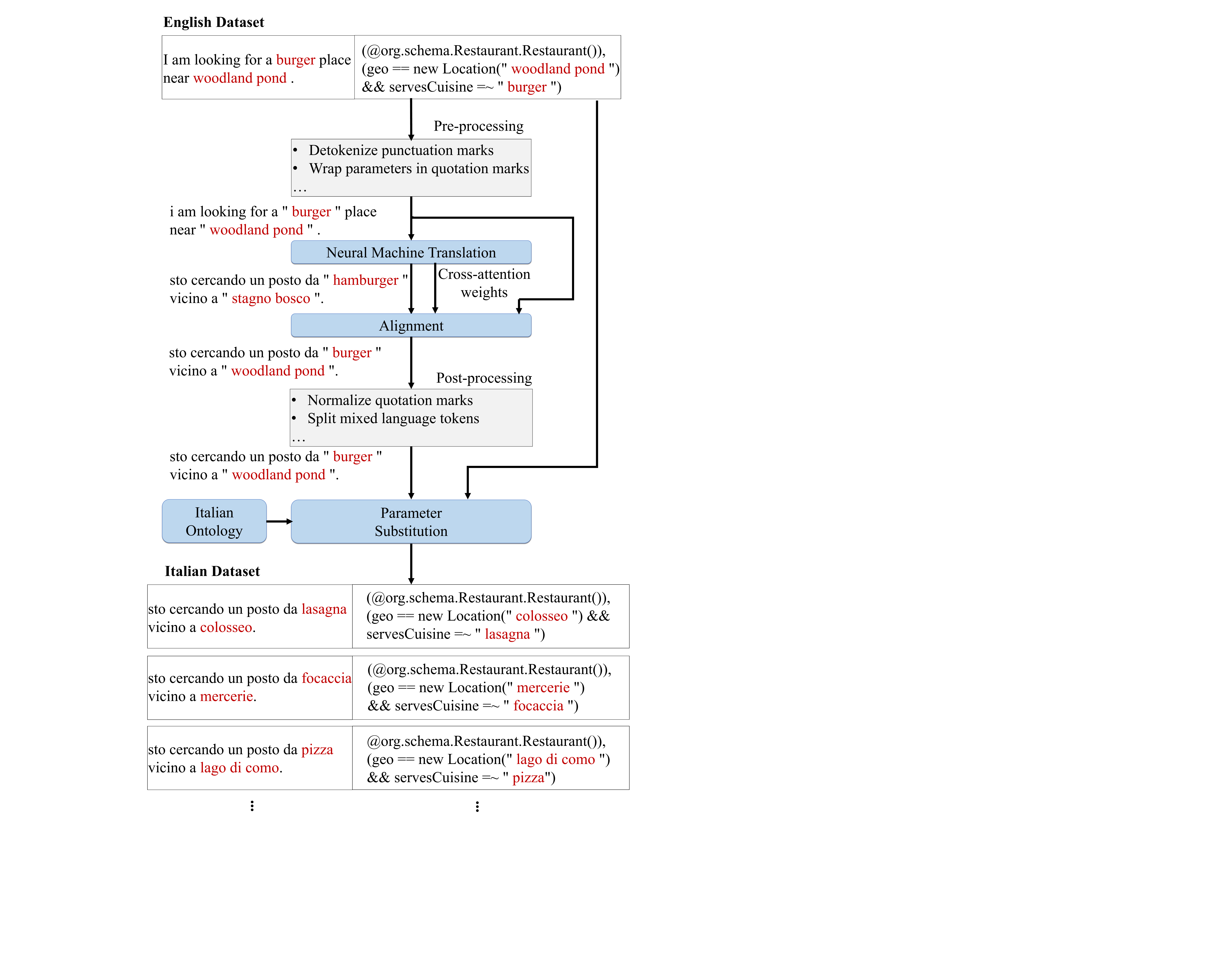}
\vspace{-0.5em}
\caption{Data generation pipeline used to produce train and validation splits in a new language such as Italian. Given an input sentence in English and its annotation in the formal ThingTalk query language~\cite{xu2020schema2qa}, SPL  generates multiple examples in the target language with localized entities.}
\vspace{-1em}
\label{fig:data_collection}
\end{figure}

For English, previous work has shown it is possible to bootstrap a semantic parser without massive amount of manual annotation, by using a large, hand-curated grammar of natural language~\cite{overnight, xu2020schema2qa}. 
This approach is expensive to replicate for all languages, due to the effort and expertise required to build such a grammar. Hence, we investigate the question: \textit{Can we leverage previous work on English semantic parsers for other languages by using machine translation?} And in particular, can we do so without requiring experts in each language?

The challenge is that a semantic parser localized to a new target language must understand questions using an ontology in the target language. For example, whereas a restaurant guide in New York may answer questions about restaurants near Times Square, the one in Italy should answer questions about restaurants near the ``Colosseo'' or ``Fontana di Trevi'' in Rome, in Italian. In addition, the parser must be able to generalize beyond a fixed set of ontology where sentences refer to entities in the target language that are unseen during training. 


We propose a methodology that leverages machine translation to localize an English semantic parser to a new target language, where the only labor required is manual translation of a few hundreds of annotated sentences to the target language. 
Our approach, shown in Fig.~\ref{fig:data_collection}, is to convert the English training data into training data in the target language, with all the parameter values in the questions and the logical forms substituted with local entities. Such data trains the parsers to answer questions about local entities. A small sample of the English questions from the evaluation set is translated by native speakers with no technical expertise, as a few-shot boost to the automatic training set. The test data is also manually translated to assess how our model will perform on real examples. We show that this approach can boost the accuracy on the English dataset as well from 64.6\% to 71.5\% for hotels, and from 68.9\% to 81.6\% for restaurants.

\begin{CJK}{UTF8}{min}
\begin{table*}[tb]
\vskip 0.1in
\fontsize{7}{9}\selectfont
\centering
\begin{tabular}{c|c|c}

\toprule

\bf{Language} & \bf{Country} & {\bf Examples} \\
\hline
\multicolumn{3}{c}{\bf Hotels} \\
\hline

English & \includegraphics[width=0.03\linewidth,height=1.2em]{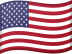} & I want a hotel near times square that has at least 1000 reviews.\\
Arabic & \includegraphics[width=0.03\linewidth,height=1.2em]{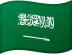} & {\fontsize{6pt}{6pt}\selectfont \< ana 'aryd fndq baalqarb men msjd al.hraam y.htwy `aly 1000 t`alyq `aly al'aql.> }  \\
German & \includegraphics[width=0.03\linewidth,height=1.2em]{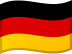}  & Ich möchte ein hotel in der nähe von marienplatz, das mindestens 1000 bewertungen hat. \\
Spanish & \includegraphics[width=0.03\linewidth,height=1.2em]{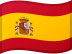} &  Busco un hotel cerca de puerto banús que tenga al menos 1000 comentarios.
 \\
Farsi & \includegraphics[width=0.03\linewidth,height=1.2em]{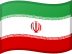} & {\fontsize{6pt}{6pt}\selectfont \<\novocalize \setfarsi  htly dr nzdyky baa.g arm my_hwaahm keh .hdaaql 1000 nqd daa^sth baa^sd. > }\\
Finnish & \includegraphics[width=0.03\linewidth,height=1.2em]{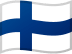}  & Haluan paikan helsingin tuomiokirkko läheltä hotellin, jolla on vähintään 1000 arvostelua.
\\
Italian & \includegraphics[width=0.03\linewidth,height=1.2em]{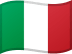} & Voglio un hotel nei pressi di colosseo che abbia almeno 1000 recensioni.
\\
Japanese & \includegraphics[width=0.03\linewidth,height=1.2em]{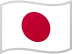} &  東京スカイツリー周辺でに1000件以上のレビューがあるホテルを見せて。
 \\
Polish & \includegraphics[width=0.03\linewidth,height=1.2em]{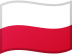}  & Potrzebuję hotelu w pobliżu zamek w malborku, który ma co najmniej 1000 ocen.
\\
Turkish & \includegraphics[width=0.03\linewidth,height=1.2em]{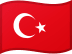} & Kapalı carşı yakınlarında en az 1000 yoruma sahip bir otel istiyorum. \\
Chinese & \includegraphics[width=0.03\linewidth,height=1.2em]{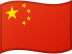} & \begin{CJK*}{UTF8}{gbsn} 我想在天安门广场附近找一家有至少1000条评论的酒店。 \end{CJK*} \\

\hline
\multicolumn{3}{c}{\bf Restaurants} \\
\hline

English & \includegraphics[width=0.03\linewidth,height=1.2em]{figures/flags/us.png} & find me a restaurant that serves burgers and is open at 14:30 . \\
Arabic & \includegraphics[width=0.03\linewidth,height=1.2em]{figures/flags/sa.png}  & {\fontsize{6pt}{6pt}\selectfont \< a`a_tr ly `aly m.t`am yqdm .t`aam al^saawrmaa w yaft.ah b.hlwl alsA`at 30:14. > } \\
German & \includegraphics[width=0.03\linewidth,height=1.2em]{figures/flags/de.png} & Finden sie bitte ein restaurant mit maultaschen essen, das um 14:30 öffnet. \\
Spanish & \includegraphics[width=0.03\linewidth,height=1.2em]{figures/flags/es.png}  &  Busque un restaurante que sirva comida paella valenciana y abra a las 14:30. \\
Farsi & \includegraphics[width=0.03\linewidth,height=1.2em]{figures/flags/ir.png} & {\fontsize{6pt}{6pt}\selectfont \<\novocalize \setfarsi rstwraany braaym peydaa kn keh ^gw^gh kebaab daa^sth baa^sd w saa`at 30:14 baaz baa^sd. > } \\
Finnish & \includegraphics[width=0.03\linewidth,height=1.2em]{figures/flags/fi.png} & Etsi minulle ravintola joka tarjoilee karjalanpiirakka ruokaa ja joka aukeaa kello 14:30 mennessä.  \\
Italian & \includegraphics[width=0.03\linewidth,height=1.2em]{figures/flags/it.png}  & Trovami un ristorante che serve cibo lasagna e apre alle 14:30. \\
Japanese & \includegraphics[width=0.03\linewidth,height=1.2em]{figures/flags/jp.png}  & 寿司フードを提供し、14:30までに開店するレストランを見つけてください。\\
Polish & \includegraphics[width=0.03\linewidth,height=1.2em]{figures/flags/pl.png} & Znajdź restaurację, w której podaje się kotlet jedzenie i którą otwierają o 14:30. \\
Turkish & \includegraphics[width=0.03\linewidth,height=1.2em]{figures/flags/tr.png} & Bana köfte yemekleri sunan ve 14:30 zamanına kadar açık olan bir restoran bul.. \\
Chinese & \includegraphics[width=0.03\linewidth,height=1.2em]{figures/flags/cn.png} & \begin{CJK*}{UTF8}{gbsn} 帮我找一家在14:30营业并供应北京烤鸭菜的餐厅。 \end{CJK*}\\

\bottomrule
\end{tabular}
\vspace{-0.5em}
\caption{Example of queries that our multilingual QA system can answer in English and 10 other languages.} 
\vspace{-1.8em}
\label{tab:sample_data}
\end{table*}
\end{CJK}


We apply our approach on the \textit{Restaurants} and \textit{Hotels} datasets introduced by \newcite{xu2020schema2qa}, which contain complex queries on data scraped from major websites. We demonstrate the efficiency of our methodology by creating neural semantic parsers for 10 languages: Arabic, German, Spanish, Persian, Finnish, Italian, Japanese, Polish, Turkish, Chinese. The models can answer complex questions about hotels and restaurants in the respective languages.
An example of a query is shown for each language and domain in Table~\ref{tab:sample_data}.

Our contributions include the following:
\begin{itemize}[leftmargin=*,noitemsep]

\item
Semantic Parser Localizer (SPL), a new methodology to localize a semantic parser for any language for which a high-quality neural machine translation (NMT) system is available. To handle an open ontology with entities in the target language, we propose {\em machine translation with alignment}, which shows the alignment of the translated language to the input language. This enables the substitution of English entities in the translated sentences with localized entities. 
Only a couple of hundred of sentences need to be translated manually;   
no manual annotation of sentences is necessary.

\item
An improved neural semantic parsing model, based on BERT-LSTM~\cite{xu2020schema2qa} but using the XLM-R encoder. Its applicability extends beyond multilingual semantic parsing task, as it can be deployed for any NLP task that can be framed as sequence-to-sequence. Pretrained models are available for download.
\item 
Experimental results of SPL for answering questions on hotels and restaurants in 10 different languages. On average, across the 10 languages, SPL achieves a logical form accuracy of 66.7\% for hotels and 71.5\% for restaurants, which is comparable to the English parser trained with English synthetic and paraphrased data. 
Our method outperforms the previous state of the art and two other strong baselines by between 30\% and 40\%, depending on the language and domain. This result confirms the importance of training with local entities.

\item 
To the best of our knowledge, ours is the first multilingual semantic parsing dataset with localized entities. Our dataset covers 10 linguistically different languages with a wide range of syntax. We hope that releasing our dataset will trigger further work in multilingual semantic parsing.

\item
SPL has been incorporated into the parser generation toolkit, Schema2QA~\cite{xu2020schema2qa}, which generates QA semantic parsers that can answer complex questions of a knowledge base automatically from its schema. With the addition of SPL, developers can easily create multilingual QA agents for new domains cost-effectively. 
\end{itemize}

\section{Related Work}
\label{related_work}

\paragraph{Multi-lingual benchmarks}
 Previous work has shown it is possible to ask non-experts to annotate large datasets for applications such as natural language inference~\cite{conneau2018xnli} and machine reading~\cite{clark2020tydi}, which has led to large cross-lingual benchmarks~\cite{hu2020xtreme}. Their approach is not suitable for semantic parsing, because it requires experts that know both the formal language and the natural language.

\paragraph{Semantic Parsing}
Semantic parsing is the task of converting natural language utterances into a formal representation of its meaning. Previous work on semantic parsing is abundant, with work dating back to the 70s~\cite{Woods1977LunarRI, 10.5555/1864519.1864543, kate2005learning, berant2013semantic}. State-of-the-art methods, based on sequence-to-sequence neural networks, require large amounts of manually annotated data~\cite{dong-lapata-2016-language, jia-liang-2016-data}. Various methods have been proposed to eliminate manually annotated data for new domains, using synthesis~\cite{overnight,DBLP:journals/corr/abs-1801-04871, geniepldi19, xu2020schema2qa, autoqa}, transfer learning~\cite{zhong2017seq2sql, decoupling2018, DBLP:journals/corr/abs-1809-08887, moradshahi2019hubert}, or a combination of both~\cite{rastogi2019towards, campagna2020zeroshot}. All these works focus mainly on the English language, and have not been applied to other languages.

\paragraph{Cross-lingual Transfer of Semantic Parsing}
\newcite{duong-etal-2017-multilingual} investigate cross-lingual transferability of knowledge from a source language to the target language by employing cross-lingual word embedding. They evaluate their approach on the English and German splits of NLmaps dataset \cite{haas-riezler-2016-corpus} and on a code-switching test set that combines English and German words in the same utterance. However, they found that joint training on English and German training data achieves competitive results compared to training multiple encoders and predicting logical form using a shared decoder. This calls for better training strategies and better use of knowledge the model can potentially learn from the dataset.

The closest work to ours is Bootstrap \cite{bootstrap}, which explores using public MT systems to generate training data for other languages. They try different training strategies and find that using a shared encoder and training on target language sentences and unmodified logical forms with English entities yields the best result. Their evaluation is done on the ATIS \cite{dahl1994expanding} and Overnight \cite{overnight} datasets, in German and Chinese. These two benchmarks have a very small number of entities. As a result, their method is unsuitable for the open ontology setting, where the semantic parser must detect entities not seen during training. 
To collect real validation and test utterances, \newcite{bootstrap} use a three-staged process to collect data from Amazon Mechanical Turkers (AMTs). They ask for three translations each per English source sentence with the hypothesis that this will collect at least one adequate translation. 
We found this approach to be less cost-effective than using professional translators. Since this process is done for the test data, it is important for the translations to be verified and have high quality. 

\section{Multi-Lingual Parser Generation}
\label{sec:methodology}

Our goal is to localize an English semantic parser for question answering that operates on an open ontology of localized entities, with no manual annotation and a limited amount of human translation. 

\subsection{Overview}

Our methodology is applicable to any semantic parser for which an English dataset is available, and for which the logical form ensures that the parameters appear exactly in the input sentence. We note that many previous techniques can be used to obtain the initial English dataset in a new domain.

Our methodology consists of the following steps:
\begin{enumerate}[leftmargin=*,noitemsep]
\item Generate training data in the target language from the English training data and an ontology of localized entities, as discussed below. 

\item 
Translate evaluation and test data in English to the target language.  To ensure that our test set is realistic, so high accuracy is indicative of good performance in practice, we engage professional translators, who are native speakers of the target language.  We ask them to provide the most natural written form of each sentence in their language, equivalent to how they would type their queries for a text-based virtual assistant. 

\item Train a semantic parser to translate sentences in the target language to the logical form using the generated sentences and a few shot of the manually translated sentences. Our semantic parsing model is described in Section~\ref{sec:nn}.
\end{enumerate}

\subsection{Training Data with Localized Entities}
\label{sec:trainingdata}
The overall architecture of our approach is illustrated in Fig.~\ref{fig:data_collection}, which shows how the English query, ``I am looking for a burger place near Woodland Pond'' is used to generate Italian training samples looking for ``lasagna'' in ``Via Del Corso'', ``focaccia'' in ``Mercerie'', and ``pizza'' in ``Lago di Como'', with the help of an open ontology of Italian entities.  Each of the sentences is annotated with their appropriate entities in the native language.  
This example illustrates why we have to handle the parameters of the queries carefully. While ``burger" is translated into ``hamburger", ``Woodland Pond'', a place in New York, is translated into ``laghetto nel bosco'', which is literally a ``pond in the woods''; these entities no longer match the entities in the target logical form. 
In general, during translation, input tokens can be modified, transliterated, omitted, or get mapped to a new token in the target language. If the semantics of the generated utterance in the target language is changed, the original logical form will no longer be the correct annotation of the utterance.

After translation, we substitute the entities with localized ones, and ensure the parameters in the sentences match those in the logical form.
To do so, we add a pair of pre- and post-processing steps to the translation to improve the outcome of the translation with public NMT models, based on error analysis.
For example, we found that the presence or absence of punctuation marks affect translation results for Persian and Arabic more than other languages. Furthermore, for languages such as Chinese and Japanese, where there is no white space delimitation between words in the sentence, the quotation marks are sometimes omitted during translation, which makes entity tracking difficult. We post-process the sentence using regular expressions to split English parameters from Chinese tokens. For Marian models, we also wrap placeholders for numbers, time, date in quotation marks to ensure they are not translated either.  

\subsection{Validation and Test Data}
As discussed above, a small amount of annotated data in the target language is translated by professional translators.  We create sentences with localized entities by showing to the translators English sentences where parameters are replaced with placeholders (numbers, dates) or wrapped in quotation marks (restaurant and hotel names, cuisine types, etc.). We ask the translators to keep the parameters intact and not translate them. The parameters are substituted later with local values in the target language. 


\section{Model Description}
\label{model}
This section first describes our translation models, then the neural semantic parser we train with the generated data. 

\subsection{Machine Translation Models}

To translate our training data, we have experimented with both pretrained Marian models ~\cite{mariannmt} and the Google public NMT system~\cite{wu2016google} (through the Google Translate API).  Marian models have an encoder-decoder architecture similar to BART~\cite{lewis2019bart} and are available in more than 500 languages and thousands of language pairs. Although Google NMT has generally higher quality than Marian models for different languages pairs and is widely adopted by different systems, Marian is preferred for two reasons. First, Marian provides flexibility, as translation is controlled and can be tuned to generate different translations for the same sentence. Second, the cost of using Google NMT to extend our work to hundreds of languages is prohibitive. 

\begin{figure}[tb]
\centering
\includegraphics[width=1\linewidth]{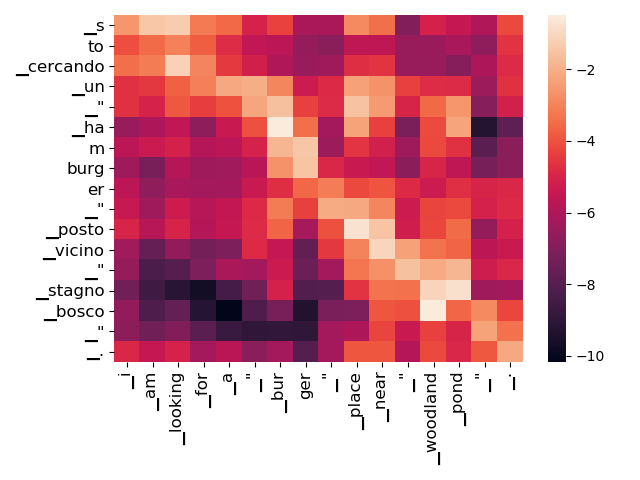}
\vspace{-2.6em}
\caption{Cross-attention weights are shown for word-pieces in the source (X axis) and target (Y axis) language. Lighter colors correspond to higher weights. The translation is different than the one in Figure~\ref{fig:data_collection} as we are using Marian instead of GT.}
\vspace{-1.5em}
\label{fig:heatmap}
\end{figure}

\subsubsection{Marian with Alignment}
\label{sec:alignment}
To find the mapping between entities in the source and the translated language, we need to 1) detect entity spans in the output sentence, 2) align those spans with input sentence spans. We have created an alignment module, which uses the cross attention weights between the encoder and the decoder of the Marian model to align the input and output sentences. These weights show the amount of attention given to each input token when an output token is being generated. 
Figure~\ref{fig:heatmap} shows a heatmap of cross-attention weights for an English sentence and its translation in Italian. The cross-attention score for each decoder token is calculated by doing a multi-head attention~\cite{vaswani2017attention} over all encoder tokens. For the example shown in Figure~\ref{fig:heatmap}, each attention vector corresponds to one column in the heatmap.

To simplify the identification of the spans, we mark each entity in the source sentence with quotation marks, using the information in the logical form. We found empirically that the quotation marks do not change the quality of the translation. When all quotation marks are retained in the translated sentences, the spans are the contiguous tokens between quotation marks in the translated sentence.  Each quotation mark in the source is aligned with the quotation mark in the target that has the highest cross-attention score between the two. If some quotations marks are not retained, however, we find the positions in the translated sentence that share the highest cross-attention score with the quotation marks surrounding each entity, to determine its span. Once spans are detected, we override target sentence spans with source sentence spans.

\subsubsection{Alignment with Google NMT}

As we cannot access the internals of Google NMT, we localize the entities by (1)
replacing parameter values in the input sentence and logical form pairs with placeholders, (2) translating the sentences, and (3) replacing the placeholders with localized entities.  Substituting with placeholders tends to degrade translation quality because the actual parameters provide a better context for translation. 

We experimented with other methods such as 1) using a glossary-based approach where parameters are detected and masked during translation and 2) replacing parameters with values from the target language before translation.  Both show poorer translation quality.  
The former technique degrades sentence quality as masking the entity reduces context information the internal transformer model relies upon to generate target sentences. The second approach creates mixed-language sentences, requiring NMT sentences to perform code-switching. It also makes the sentences look less natural and shifts input distribution away from what public NMTs have been trained on. 

\subsection{Semantic Parsing Model}
\label{sec:nn}
The neural semantic parser we train using our translated training data is based on the previously proposed BERT-LSTM architecture~\cite{xu2020schema2qa}, which we modify to use the XLM-R pretrained model~\cite{conneau2019unsupervised} as the encoder instead.
Our model is an encoder-decoder neural network that uses the XLM-R model as an encoder and a LSTM decoder with attention and pointer-generator~\cite{see2017get}. More details are provided in Appendix~\ref{sp_model}. As in previous work~\cite{xu2020schema2qa}, we apply rule-based preprocessing to identify times, dates, phone numbers, etc. All other tokens are lower cases and split into subwords according to the pretrained vocabulary. The same subword preprocessing is applied to entity names that are present in the output logical form.

\begin{table*}[htb]
\vskip 0.1in
\centering
\scalebox{0.7}{
\begin{tabular}{l|r|r|r|r|r}
\toprule
\multirow{3}{*}{\bf Metrics} & \multicolumn{5}{c}{\bf Dataset}  \\
\hline
\bf{}  & {\bf Overnight (Blocks)}  & {\bf Overnight (Social)} & {\bf  ATIS} & {\bf Schema2QA (Hotels)} & {\bf Schema2QA (Restaurants)} \\
\hline
\# attributes & 10 & 15 & 16 & 18 & 25  \\
\# examples & 1,305 & 2,842  & 4,433 & 363,101 & 508,101  \\
avg \# unique unigrams per example   & 6.82 & 8.65 & 7.75  & 12.19 & 12.16  \\
avg \# unique bigrams per example  & 7.44 & 8.68 & 6.99  & 11.62 & 11.57 \\
avg \# unique trigrams per example   & 6.70  & 7.90 & 6.03 & 10.64 & 10.59  \\
avg \# properties per example   & 1.94 & 1.65  & 2.56 & 2.03 & 2.06 \\
avg \# values per property  & $\leq2$ & $\leq2$ & $\leq20$  & $\geq100$ & $\geq100$   \\

\bottomrule
\end{tabular}
}
\vspace{-0.5em}
\caption{Statistical analysis of the training set for Overnight, ATIS, and schema2QA datasets. For overnight, the two domains with the lowest reported accuracies are chosen.}
\vspace{-1.5em}
\label{tab:train_stats}
\end{table*}
\section{Experiments}
\label{experiments}

We have implemented the full SPL methodology in the form of a tool. Developers can use the SPL tool to create a new dataset and semantic parser for their task.
We evaluate our models on the Schema2QA dataset~\cite{xu2020schema2qa}, translated to other languages using our tool. We first describe our dataset and then show our tool's accuracy, both without any human-produced training data (zero-shot) and if a small amount of human-created data in the target language is available (few-shot). In our experiments, we measure the \textit{logical form exact match} (em) accuracy, which considers the result to be correct only if the output matches the gold logical form token by token.
We additionally measure the \textit{structure match} (sm) accuracy, which measures whether the gold and predicted logical forms are identical, ignoring the parameter values. 
A large difference between exact and structure accuracy indicates that the parameters are poorly handled. 
We report results on both validation and test sets. 
We present the results for both restaurants and hotels domain in this paper.  

Our toolkit uses the Genie~\cite{geniepldi19} library for synthesis and data augmentation. Our models were implemented using the Huggingface~\cite{Wolf2019HuggingFacesTS} and GenieNLP\footnote{\footnotesize{\url{https://github.com/stanford-oval/genienlp}}} libraries.







\subsection{Dataset}
Using our approach, we have constructed a multilingual dataset based on the previously proposed Schema2QA \textit{Restaurants} and \textit{Hotels} datasets~\cite{xu2020schema2qa}. These datasets contain questions over scraped Schema.org web data, expressed using the ThingTalk query language. ThingTalk is a subset of SQL in expressiveness, but it is more tailored to natural language translation. The training sets are constructed using template-based synthesis and crowdsourced paraphrasing, while the validation and test sets are crowdsourced and manually annotated by an expert.  

We chose these two datasets as a starting point as they require understanding both complex questions and a large number of entities, many of which are not seen in training. Note that the parameters in the logical forms are \textit{aligned} with those in the input utterance: every open-ontology parameter value must appear exactly in the utterance. Table~\ref{tab:train_stats} shows the comparison of our dataset with the  Overnight~\cite{overnight} and ATIS datasets~\cite{dahl1994expanding}, which previous work has translated to other languages~\cite{bootstrap}. The Schema2QA dataset is larger, has more linguistic variety, and has significantly more possible values for each property.
The hotels domain contains 443 and 528 examples, and the restaurants domain contains 528 and 524 examples in the validation and test splits, respectively. 

We scrape Yelp, OpenTable, and TripAdvisor websites for localized ontologies on restaurants, and Hotels.com and TripAdvisor for hotels.  To ensure that some entities are unseen in validation and test, each ontology is split into two, for (1) training, and (2) validation and test.  The two splits overlap between 40\% to 50\%, depending on the domain and language. We replace the English parameters in the translated sentences with localized entities.

We have translated the Schema2QA dataset to 10 different languages, chosen to be linguistically diverse. 
To translate the training set, we use Google NMT for Farsi, Japanese, and Turkish. We use Marian for the other seven languages. Marian BLEU scores for all language pairs are available online\footnote{\footnotesize{\url{https://github.com/Helsinki-NLP/Tatoeba-Challenge/tree/master/results}}}. In our initial experiments, we found that some of the models with high reported BLEU scores, such as English to Japanese, do not produce correct translations for our dataset. Thus, we perform the following to verify each model's quality: First, we choose a subset of the English evaluation set and translate it with the corresponding Marian model. The results are then back-translated to English using Google NMT. If the meaning is not preserved for at least 90\% of the samples, we use Google NMT for that language.

We chose a subset of the validation set (75\% for hotels and 72\% for restaurants) to be professionally translated. We use this data to train our parser in a few-shot setting (Section~\ref{sec:fewshot}). The full test sets for both domains are professionally translated.

\subsection{BackTranslation: Translate at Test Time}
As our first baseline, we train an English semantic parser on the English training set; at test time, the sentence (including its entities) is translated on-the-fly from the target language to English and passed to the semantic parser. 

The experimental results are shown in Table~\ref{tab:direct_hotels}.
The results vary from a minimum of 9.7\% for Japanese to a maximum of 34.4\% for Turkish.  
Comparing the results to English, we observe about 30\% to 50\% drop in exact match accuracy. In general, the closer the language is to English in terms of semantics and syntax, the higher the BLEU score will be using NMT. The large difference between em and sm accuracies is caused by the wrong prediction of parameter values. This is expected since the entities translated to English no longer match with the annotations containing localized entities. Note that the English parser has learned to primarily copy those parameters from the sentence.

\begin{table*}[htb]
\fontsize{8}{10}\selectfont
\centering
\begin{tabular}{l||>{\centering}p{5mm}|>{\centering}p{5mm}||c|c||c|c||c|c||c|c}

\toprule
\multirow{3}{*}{\bf Language } & \multicolumn{2}{c||}{\bf BT}  & \multicolumn{4}{c||}{\bf \experimentI} & \multicolumn{4}{c}{\bf \experimentII}  \\
\cline{2-11}
& \multicolumn{2}{c||}{\bf Test} &  \multicolumn{2}{c||}{\bf Dev} & \multicolumn{2}{c||}{\bf Test} &  \multicolumn{2}{c||}{\bf Dev} & \multicolumn{2}{c}{\bf Test} \\
\cline{2-11}
  & em & sm & em & sm & em & sm & em & sm  &em & sm \\
\hline
\multicolumn{11}{c}{\bf Hotels} \\
\hline

German & 30.4 & 50.2 & 30.2 & 55.1 & 26.9 & 50.4 & 35.7 & 56.4 & \bf 30.9 & 51.9 \\
Farsi & 19.1 & 36.0 & 23.6 & 44.2 & \bf 22.6 & 43.8 & 22.2 & 44.2 & 21.4 & 42.8 \\
Finnish & 29.2 & 46.6 & 27.8 & 53.4 & 29.1 & 52.5 & 26.9 & 51.4 & \bf 30.2 & 57.2 \\
Japanese & 15.7 & 25.2 & 20.1 & 29.4 & 19.9 & 27.3 & 21.0 & 39.3 & \bf 20.5 & 39.2 \\
Turkish & \bf 32.0 & 49.4 & 18.5 & 43.6 & 17.2 & 40.9 & 22.6 & 62.3 & 25.6 & 60.2 \\
\hline
\multicolumn{11}{c}{\bf Restaurants} \\
\hline

German & 26.9 & 51.9 & 31.3 & 62.9 & 25.8 & 60.1 & 30.3 & 61.0 & \bf 27.3 & 56.9 \\
Farsi & 14.5 & 40.8 & 20.1 & 28.2 & 15.1 & 26.2 & 22.0 & 34.1 & \bf 15.5 & 28.1 \\
Finnish & \bf 24.0 & 55.0 & 19.9 & 51.9 & 18.9 & 54.4 & 22.3 & 53.8 & 20.8 & 53.2 \\
Japanese & 9.7 & 29.8 & 19.2 & 22.1 & 18.3 & 20.6 & 22.1 & 31.6 & \bf 18.1 & 24.3 \\
Turkish & \bf 34.4 & 62.6 & 29.7 & 54.2 & 28.4 & 44.1 & 33.5 & 54.7 & 28.1 & 45.6 \\

\bottomrule
\end{tabular}
\vspace{-0.7em}
\caption{Experiment results for hotels (top rows) and restaurants (bottom rows) domain using Bootstrap and BackTranslation methods as our baseline. em and sm indicate exact and structure match accuracy respectively. We chose 5 representative languages for these experiments. Exact match accuracies for the English Test set are 64.6\% for hotels, and 68.9\% for restaurants.}
\vspace{-1.6em}
\label{tab:direct_hotels}
\end{table*}

\subsection{Bootstrap: Train with Translated Data}
As proposed by \newcite{bootstrap}, we create a new training set by using NMT to directly translate the English sentences into the target language; the logical forms containing English entities are left unmodified. This data is then used to train a semantic parser.
The results are shown in Table~\ref{tab:direct_hotels}, in the ``Bootstrap'' column. Overall, the performance of Bootstrap is comparable to the performance of BackTranslation, ranging from 15\% on Farsi restaurants to 29\% on Finnish hotels.

In a second experiment, we train a semantic parser on a dataset containing both English and translated sentences. Note that the test set is the same and contains only questions in the target language.
Training with a mixture of languages has shown improvements over single language training ~\cite{liu2020multilingual, arivazhagan2019massively}.
This experiment (shown as \experimentII in Table~\ref{tab:direct_hotels}) achieves between 16\% to 31\% accuracy outperforming \experimentVdirect for all 5 languages except for Turkish hotels and Turkish and Finnish restaurants.

Overall, these two experiments show that training with translated data can improve over translation at test time, although not by much. Furthermore, as we cannot identify the original parameters in the translated sentences, we cannot augment the training data with localized entities. This step is much needed for the neural model to generalize beyond the fixed set of values it has seen during training.
A neural semantic parser trained with Bootstrap learns to translate (or transliterate) the entity names from the foreign language representation in the sentence to the English representation in the logical form. Hence, it cannot predict the localized entities contained in the test set, which are represented in the target language.

\subsection{SPL: Semantic Parser Localizer}
There are three key components in SPL methodology: 1) Translation with alignment to ensure parameters are preserved, 2) Training with parameter-augmented machine-translated data, and 3) Boosting accuracy by adding human-translated examples to the training data simulating a few-shot setting.
Here we describe the experiments we designed to evaluate each component separately.

\begin{table*}[htb]
\fontsize{8}{10}\selectfont
\begin{tabular}{l|>{\centering}p{5mm}|>{\centering}p{5mm}||c|c||c|c||c|c||c|c||c|c||c|c}

\toprule
\multirow{3}{*}{\bf Language} & \multicolumn{2}{c||}{\bf BT (+Align.)} &  \multicolumn{4}{c||}{\bf  Zero-Shot SPL} & \multicolumn{4}{c||}{\bf Zero-Shot SPL (+English) }&  \multicolumn{4}{c}{\bf Few-Shot SPL}   \\
\cline{2-15}
& \multicolumn{2}{c||}{\bf Test} &  \multicolumn{2}{c||}{\bf Dev} & \multicolumn{2}{c||}{\bf Test} &  \multicolumn{2}{c||}{\bf Dev} & \multicolumn{2}{c||}{\bf Test} &  \multicolumn{2}{c||}{\bf Dev} & \multicolumn{2}{c}{\bf Test} \\
\cline{2-15}
 & em  & sm & em & sm & em & sm & em & sm & em & sm & em & sm & em & sm   \\
\hline
\multicolumn{15}{c}{\bf Hotels} \\
\hline
Arabic & 22.7 & 26.1 & 51.6 & 59.8 & 51.3 & 60.8 & 53.8 & 61.7 & 54.9 & 61.7 & 60.8 & 67.2 & \bf 61.4 & 67.0 \\
German & 51.3 & 54.2 & 70.7 & 73.6 & 61.0 & 66.1 & 70.0 & 73.3 & 65.9 & 68.0 & 77.1 & 80.2 & \bf 68.6 & 71.2 \\
Spanish & 53.4 & 55.3 & 69.0 & 72.3 & 61.6 & 68.2 & 73.1 & 76.2 & 65.3 & 70.8 & 76.3 & 80.9 & \bf 67.0 & 72.0 \\
Farsi & 51.4 & 53.1 & 63.5 & 65.0 & 58.9 & 61.0 & 70.3 & 71.7 & 58.9 & 61.8 & 77.0 & 78.9 & \bf 63.5 & 65.4 \\
Finnish & 50.8 & 54.4 & 57.9 & 59.1 & 62.5 & 66.7 & 64.2 & 65.4 & 60.3 & 65.0 & 69.3 & 70.7 & \bf 68.4 & 71.3 \\
Italian & 53.4 & 56.8 & 66.9 & 72.6 & 60.8 & 66.7 & 66.4 & 71.2 & 64.4 & 69.9 & 69.8 & 75.8 & \bf 65.7 & 71.8 \\
Japanese & 42.3 & 44.6 & 71.0 & 72.0 & 63.6 & 65.0 & 71.3 & 72.1 & 59.5 & 61.4 & 73.1 & 78.2 & \bf 67.6 & 69.3 \\
Polish & 49.8 & 52.3 & 58.7 & 62.1 & 54.9 & 59.3 & 60.0 & 63.4 & 57.6 & 60.6 & 67.7 & 71.6 & \bf 64.8 & 68.4 \\
Turkish & 55.7 & 59.5 & 69.0 & 72.5 & 60.2 & 69.1 & 73.0 & 76.9 & 64.0 & 73.3 & 77.8 & 79.6 & \bf 69.3 & 74.4 \\
Chinese & 29.2 & 32.4 & 55.9 & 60.4 & 52.8 & 58.1 & 54.6 & 59.8 & 51.1 & 56.1 & 56.7 & 67.2 & \bf 62.9 & 67.4 \\
\midrule
\multicolumn{15}{c}{\bf Restaurants} \\
\midrule
Arabic & 34.6 & 36.1 & 66.7 & 69.0 & 67.0 & 70.0 & 60.8 & 63.7 & 67.7 & 71.6 & 75.9 & 77.4 & \bf 74.6 & 79.1 \\
German & 52.3 & 55.7 & 69.4 & 71.9 & 63.0 & 65.6 & 74.4 & 76.3 & 65.3 & 68.9 & 82.6 & 84.8 & \bf 77.1 & 80.7 \\
Spanish & 58.2 & 61.3 & 68.6 & 72.1 & 67.6 & 74.0 & 70.7 & 75.0 & 67.4 & 75.2 & 82.1 & 84.7 & \bf 77.5 & 80.5 \\
Farsi & 57.8 & 62.2 & 63.0 & 64.5 & 61.8 & 62.4 & 69.0 & 70.0 & 65.5 & 66.2 & 78.0 & 78.5 & \bf 74.2 & 75.0 \\
Finnish & 53.8 & 57.1 & 63.0 & 65.4 & 58.6 & 60.3 & 63.4 & 65.1 & 59.2 & 60.5 & 72.9 & 74.9 & \bf 68.1 & 69.7 \\
Italian & 56.1 & 59.5 & 52.1 & 53.3 & 48.3 & 50.6 & 53.3 & 54.6 & 52.9 & 55.3 & 70.3 & 72.0 & \bf 69.0 & 70.5 \\
Japanese & 49.6 & 52.5 & 45.1 & 47.0 & 41.3 & 43.6 & 48.9 & 51.1 & 48.7 & 50.5 & 75.2 & 76.5 & \bf 70.5 & 72.2 \\
Polish & 49.6 & 54.0 & 50.9 & 52.7 & 51.5 & 52.7 & 55.7 & 60.8 & 56.5 & 60.7 & 65.3 & 66.1 & \bf 64.3 & 65.1 \\
Turkish & 57.8 & 61.6 & 59.6 & 61.3 & 57.8 & 60.1 & 58.7 & 60.3 & 56.1 & 58.6 & 80.3 & 81.3 & \bf 74.6 & 76.5 \\
Chinese & 42.8 & 45.5 & 56.6 & 58.8 & 46.2 & 51.1 & 64.1 & 65.6 & 57.3 & 61.6 & 69.8 & 72.1 & \bf 65.3 & 69.7 \\

\bottomrule
\end{tabular}
\vspace{-0.7em}
\caption{Experiment results for hotels (top rows) and restaurants (bottom rows) domain using SPL. em and sm indicate exact and structure match accuracy respectively. Zero-shot and few-shot exact match accuracies for English test set are 64.6\% and 71.5\% for hotels, and 68.9\% and 81.6\% for restaurants.}
\vspace{-1em}
\label{tab:spl_results}
\end{table*}

\subsubsection{Test Time Translation with Alignment}
In this experiment, we run BackTranslation (BT) with alignment to understand its effect.  
We translate sentences from the foreign language to English at test time, but we use the entity aligner described in Section~\ref{sec:alignment} to copy the localized entities in the foreign language to the translated sentence before feeding it into the English semantic parser.
The results, as shown in Table~\ref{tab:spl_results}, improve by 25\% to 40\% across all languages compared to naive BT. This highlights the importance of having entities that are aligned in the sentence and the logical form, as that enables the semantic parser to copy entities from the localized ontology for correct prediction. This is evident as the exact accuracy result is close to that of structure accuracy.  

\subsubsection{Training with Machine Translated Data}

In the next experiment, we apply the methodology in Section~\ref{sec:methodology} to the English dataset to create localized training data and train one semantic parser per language. We translate a portion of the validation set using human translators and combine it with the machine-translated validation data.
For all the following experiments, the model with the highest em accuracy on this set is chosen and tested on human-translated test data. 

As shown in Table~\ref{tab:spl_results}, the results obtained by this methodology outperforms all the baselines. Specifically, we achieve improvements between 33\% to 50\% over the previous state-of-the-art result, represented by the Bootstrap approach. The neural model trained on SPL data takes advantage of entity alignment in the utterance and logical form and can copy the entities directly. The exact match accuracy ranges from 53\% in Chinese to 62\% in Spanish for hotels, and from 41\% in Japanese to 68\% in Spanish for restaurants. 
Comparing to the accuracy of 65\% and 69\% for hotels and restaurants in English, respectively, we see a degradation in performance for languages that are very different from English.  Languages close to English, such as Spanish, approach the performance of English. 

\subsubsection{Adding English Training Data}
Similar to Bootstrap (+English), we also experimented with combining the original English training set with the training set generated using SPL approach. 
Except for some drops (0.3\%-4\%) in accuracy for Spanish and Turkish restaurants and Finnish and Japanese hotels, we observe about 1\% to 10\% improvement compared to when English training data is not used. 
As the parser is exposed to a larger vocabulary and two potentially different grammars at once, it must learn to pay more attention to sentence semantics as opposed to individual tokens.
Additionally, the English training data contains human-paraphrased sentences, which are more natural compared to synthetic data, and add variety to the training set. 



\subsubsection{Adding a Few Human Translation to Training Data}
\label{sec:fewshot}
In our final experiment, we add the portion of the validation set translated by humans to the training set generated using SPL. Since the validation size is much smaller than the training size (0.03\% for hotels and 0.12\% for restaurants), this is similar to a few-shot scenario where a small dataset from the test distribution is used for training. 

In Table~\ref{tab:hans} we have computed BLEU~\cite{papineni2002bleu} and TER~\cite{snover2006study} scores between machine-translated and human-translated validation data for hotels. One key takeaway is that machine-translated data has quite a different distribution than human-translated data as none of the BLUE scores are higher than 0.45. Adding a few real examples can shrink this gap and yield higher accuracy on natural sentences. 

As shown in Table~\ref{tab:spl_results}, the test results are improved significantly across all the languages for both domains.
This shows that a small addition of real training data improves the model performance significantly. The exact match accuracy varies across languages, with a low of 61.4\% on Arabic and a high of 69.3\% on Turkish for hotels, and a low of 64.3\% on Polish and a high of 77.1\% on Spanish for restaurants.  The multilingual results compare favorably with those for English.  
We show that a few-shot boost of crowdsourced evaluation data in training can also improve the English semantic parser, raising its accuracy from 65\% to 72\% for hotels, and from 69\% to 82\% for restaurants.  
The few-shot approach is particularly helpful when the training and test data are collected using different methods; this can create a new avenue for further research on multilingual tasks.

We have performed an error analysis on the results generated by the parser. At a high level, we found the biggest challenge is in recognizing entities, in particular, when entities are unseen, and when the type of the entities is ambiguous. We also found translation noise would introduce confusion for implicit concepts such as ``here''. Translation sometimes introduces or removes these concepts from the sentence. Detailed error analysis is provided in Appendix~\ref{sec:erroranalysis}.

\section{Conclusion}
\label{conclusion}


This paper presents SPL, a toolkit and methodology to extend and localize semantic parsers to a new language with higher accuracy, yet at a fraction of the cost compared to previous methods. SPL was incorporated into the Schema2QA toolkit to give it a multilingual capability.

SPL can be used by any developer to extend their QA system's current capabilities to a new language in less than 24 hours, leveraging professional services to translate the validation data and mature public NMT systems. We found our approach to be effective on a recently proposed QA semantic parsing dataset, which is significantly more challenging than other available multilingual datasets in terms of sentence complexity and ontology size.

Our generated datasets are automatically annotated using logical forms containing localized entities; we require no human annotations. Our model outperforms the previous state-of-the-art methodology by between 30\% and 40\% depending on the domain and the language. Our new datasets and resources are released open-source\footnote{\footnotesize{\url{https://github.com/stanford-oval/SPL}}}. Our methodology enables further investigation and creation of new benchmarks to trigger more research on this topic.

\section*{Acknowledgements}
We would like to thank Sierra Kaplan-Nelson and Max Farr for their help with error analysis. This work is supported in part by the National Science Foundation under Grant No.~1900638 and the Alfred P. Sloan Foundation under Grant No.~G-2020-13938.

\bibliographystyle{acl_natbib}
\bibliography{anthology,emnlp2020}

\appendix
\begin{table*}[htb]
\vskip 0.1in
\fontsize{8}{10}\selectfont
\centering
\begin{tabular}{l|c|c|c|c|c}

\toprule
\bf{Language} & BLEU score (\%) & GLEU score (\%) & METEOR score (\%) & NIST score (\%) & TER score (\%)  \\
\hline

Arabic & 15.9 & 19.1 & 32.9 & 49.7 & 72.4 \\ 
German & 37.2 & 41.8 & 61.3 & 45.3 & 45.5 \\ 
Spanish & 42.0 & 46.8 & 64.9 & 43.1 & 39.2 \\ 
Farsi & 20.9 & 25.9 & 44.5 & 44.4 & 63.5 \\ 
Finnish & 23.7 & 27.9 & 43.0 & 34.3 & 52.7 \\ 
Italian & 32.7 & 38.5 & 56.9 & 49.8 & 44.2 \\ 
Japanese & 44.0 & 46.6 & 67.9 & 22.7 & 53.8 \\ 
Polish & 18.5 & 22.2 & 38.1 & 48.3 & 62.4 \\ 
Turkish & 26.2 & 30.2 & 47.0 & 43.7 & 52.8 \\ 
Chinese & 21.7 & 26.1 & 42.8 & 32.9 & 76.9 \\ 

\midrule
\midrule

Arabic & 21.3 & 23.7 & 36.6 & 48.8 & 65.7 \\ 
German & 35.2 & 39.6 & 57.8 & 39.1 & 48.0 \\ 
Spanish & 41.8 & 46.2 & 61.2 & 38.7 & 42.6 \\ 
Farsi & 23.8 & 28.3 & 46.6 & 45.9 & 59.9 \\ 
Finnish & 28.0 & 33.5 & 47.4 & 34.6 & 46.6 \\ 
Italian & 3.07 & 42.6 & 57.6 & 43.4 & 43.1 \\ 
Japanese & 44.2 & 46.4 & 68.0 & 23.1 & 58.5 \\ 
Polish & 23.6 & 28.1 & 41.0 & 35.0 & 56.8 \\ 
Turkish & 28.2 & 32.4 & 48.0 & 31.7 & 48.3 \\ 
Chinese & 25.6 & 31.1 & 50.9 & 34.8 & 59.7 \\


\bottomrule
\end{tabular}
\caption{Results for different similarity metrics. The results are shown for the hotels validation set.}
\label{tab:hans}
\end{table*}

\section{Semantic Parser Model}
\label{sp_model}

Our neural semantic parser is based on BERT-LSTM model~\cite{xu2020schema2qa}, a previously-proposed model that was found effective on semantic parsing. 
We have modified the model encoder to use XLM-R instead of BERT or LSTM encoder. XLM-R is a Transfomer-based multilingual model trained on CommonCrawl corpus in 100 different languages. Unlike some XLM~\cite{lample2019crosslingual} multilingual models, it can detect the language from the input ids without requiring additional language-specific tokens.
The decoder is an LSTM decoder with attention and a pointer-generator. At each decoding step, the model decides whether to generate a token or copy one from the input context.

We preprocess the input sentences by lowercasing all tokens except for entity placeholders such as TIME\_0, DATE\_0, etc. and splitting tokens on white space. 
The formal code tokens are also split on whitespace, but their casing is preserved.
XLM-R uses the sentence piece model to tokenize input words into sub-word pieces. For the decoder, to be able to copy tokens from pretrained XLM-R vocabulary,  we perform the same sub-word tokenization of parameter values in the input sentence and in the formal language.

The word-pieces are then numericalized using an embedding matrix and fed into a 12-layer pretrained transformer network which outputs contextual representations of each sub-word. The representations are then aggregated using a pooling layer which calculates the final representation of the input sentence:

\begin{equation*}
    \begin{aligned}
    \label{eq:eq1}
        H &= \mW_{agg}\text{ReLU}(\mW_{E}\text{mean}(h^{0}, h^{1}, ..., h^{N}))
    \end{aligned}
\end{equation*}

\noindent where $H$ is the final sentence embedding, $\mW_{agg}$ and $\mW_{E}$ are learnable weights, $\text{relu}(.)$ is the rectified linear unit function, and $\text{mean}(.)$ is the average function.\footnote{Bias parameters are omitted for brevity.}

\begin{figure}[htb]
\centering
\includegraphics[width=1\linewidth, height=5.5cm]{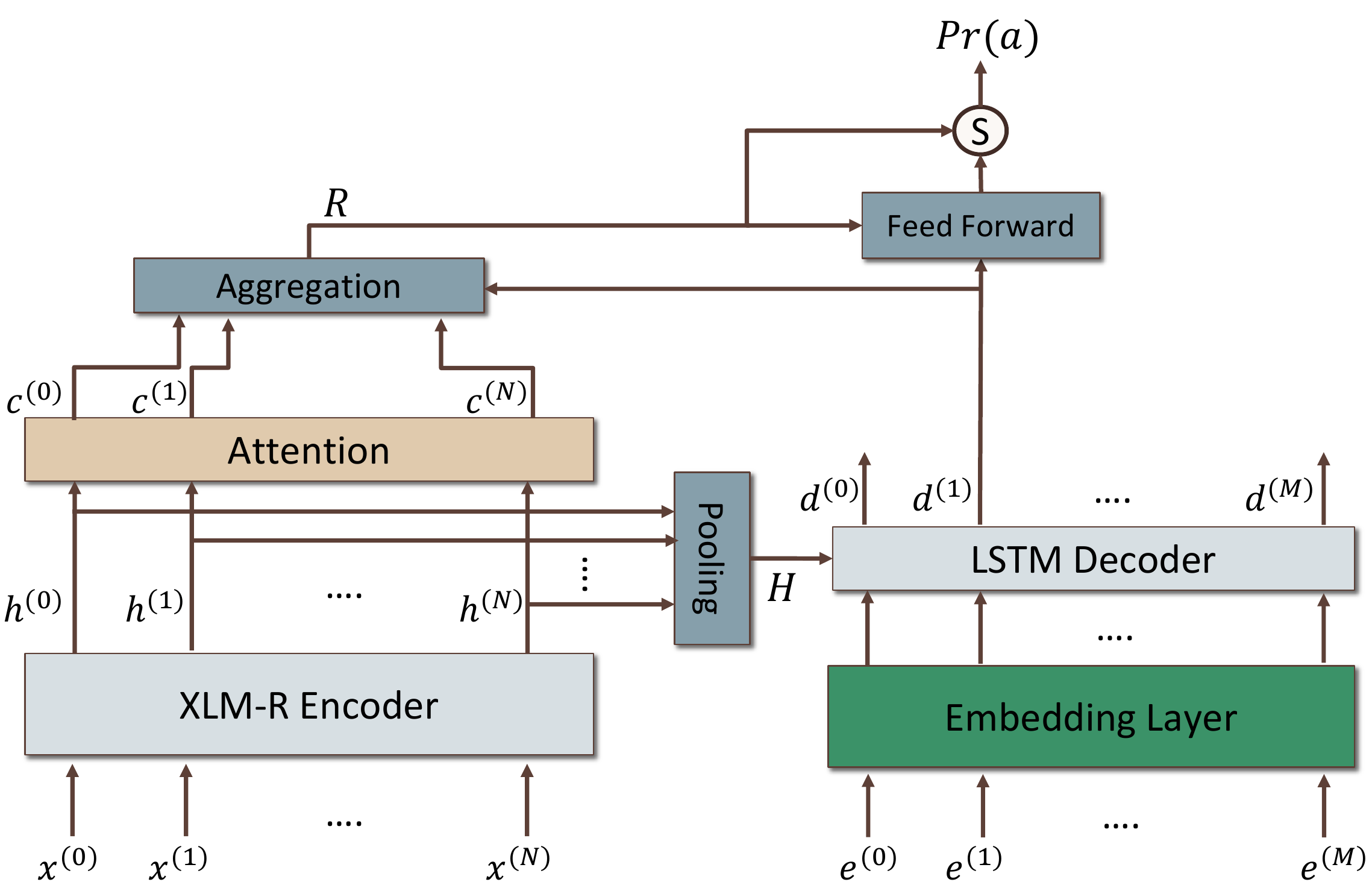}
\caption{Semantic parser neural model. It has a Seq2Seq architecture with XLM-R encoder and LSTM decoder with attention.} 
\label{fig:model}
\end{figure}

The decoder uses an attention-based pointer-generator to predict the target logical form one token at a time.
The tokenized code word-pieces are passed through a randomly initialized embedding layer, which will be learned from scratch. Using pretrained language models instead, did not prove to be useful as none of them are trained on formal languages. Each embedded value is then passed to an LSTM cell. The output is used to calculate the attention scores against each token representation from the encoder ($c^{t}$) and produce the final attention context vector ($C$). The model then produces two vocabulary distributions: one over the input sentence ($P_c(a^{t})$), and one over XLM-R sentence piece model's vocabulary ($P_v(a^{t})$). A trainable scalar switch ($s$) is used to calculate the weighted sum of the two distributions. The final output is the token with the highest probability.

\begin{equation*}
    \label{eq:eq2}
    \begin{aligned}
    &P_c(a^{t}|C, d^{t}) = \sum_{a^{t}=a^{*t}}\softmax(d^{t}C^\top)\\
    &P_v(a^{t}|C, d^{t}) = \softmax(\mW_{o}C)\\
    &P(a^{t}|C, d^{t}) = s^{t}P_c(a^{t}) + (1-s^{t})P_v(a^{t})
    \end{aligned}
\end{equation*}


The model is trained autoregressively using teacher forcing, with token-level cross-entropy loss:

\begin{equation*}
   \label{eq:eq3}
    \begin{aligned}
    \mathcal{L} = - \sum_{t=0}^{N}\sum_{a^{t}\in V} {\1[a^{t}=a^{*t}]\log P(a^{t}|C, d^{t})}
    \end{aligned}
\end{equation*}

Here $\mathcal{L}$ indicates the loss value, and $\1[.]$ is the indicator function: it is 1 when the predicted token $a$ matches the gold answer token $a^*$, and 0 otherwise.

\section{Implementation Details}

Our code implementations are in PyTorch\footnote{\footnotesize{\url{https://pytorch.org/}}} and based on HuggingFace~\cite{Wolf2019HuggingFacesTS}. In all of our experiments, we used \texttt{xlmr-base} model which is trained on CommonCrawl data in 100 languages with a shared vocabulary size of 250K. The model architecture is similar to BERT and has 12 Transformer Encoder layers with 12 attention heads each and a hidden layer dimension of 768. XLM-R uses sentence-piece model to tokenize the input sentences. We used Adam~\cite{kingma2014adam} as our optimizer
with a learning rate of $1 \times 10^{-4}$ and used transformer non-linear warm-up schedule~\cite{popel2018training}. In all our experiments we used the same value for hidden dimension (768), transformer model dimension (768), the number of transformer heads (12), size of trainable dimensions in decoder embedding matrix (50), and the number of RNN layers for the decoder (1). These parameters were chosen from the best performing model over the English dev set for each domain. Each model has a different number of parameters depending on the language trained on and the number of added vocabulary from the training and validation set. However, this number does not vary much, and the average across languages is about 300M including XLM-R parameters. We batch sentences based on their token count. We set the total number of tokens to be 5K, which would be about 400 examples per batch.
Our models were trained on NVIDIA V100 GPU using AWS platform. Single language models were trained for 60K iterations, which takes about 6 hours. For a fair comparison, models trained jointly on English and the target language were trained for 80K iterations.




\section{Error Analysis}
\label{sec:erroranalysis}
We present an error analysis for 5 languages (Spanish, Persian, Italian, Japanese, and Chinese) for which we have access to native speakers. 

\begin{itemize}[leftmargin=*,noitemsep]
  \item Locations are sometimes parsed incorrectly. In many cases, the model struggles to distinguish an explicit mention of ``here'' from no mention at all. We suspect this is due to translation noise introducing or omitting a reference to the current location.
  \item In some examples, the review author’s name is being parsed as a location name. The copying mechanism deployed by the neural model decoder relies on the context of the sentence to identify \textit{both} the type and span of the parameter values. Thus if localization is done poorly, the model will not be able to generalize beyond a fixed ontology.
  \item Occasionally, the parser has difficulty distinguishing between rating value and the number of reviews, especially if the original sentence makes no mention of \textit{starts} or \textit{posts} and instead uses more implicit terms like \textit{top} or \textit{best}.
  \item In some examples, the input sentence asks for information about ``this restaurant" but the program uses the user's home location instead of their current location.
  \item There are human mistranslations where check-in time has been mislabeled as check-out time. Additionally, sentence ambiguity is exacerbated by the human translation step, for example, between a hotel's official star rating value and the customer's average rating value. In English, this kind of ambiguity is resolved by expert annotation flagging ambiguous sentences.
  \item Translation noise in some cases, can change the numbers in the sentence. For example, ``at least" / ``more than" are equivalent in DBTalk language, but it's possible that when the translation occurs the number is changed (``at least 4" $\rightarrow$ ``more than 3").
  \item In morphologically-rich languages (such as Italian), the entities often are not in grammatical agreement with the rest of the sentence (e.g. a feminine article precedes a masculine entity), which confuses the model on the boundaries of the entity.
\end{itemize}

\end{document}